\documentclass{article}

\usepackage[final]{neurips_2024}

\usepackage{natbib}

\usepackage{multirow}
\usepackage[utf8]{inputenc} 
\usepackage[T1]{fontenc}    
\usepackage{hyperref}       
\usepackage{url}            
\usepackage{booktabs}       
\usepackage{amsfonts}       
\usepackage{nicefrac}       
\usepackage{microtype}      
\usepackage{xcolor}         
\usepackage{amsmath}
\usepackage{algorithm}
\usepackage{graphicx}
\usepackage{algpseudocode}

\newcommand{\X}{\mathcal{X}}

\title{RED – Robust Environmental Design}

\author{%
  Jinghan Yang \\
}

\begin{document}
\maketitle
\begin{abstract}
The classification of road signs by autonomous systems, especially those reliant on visual inputs, is highly susceptible to adversarial attacks. 
Traditional approaches to mitigating such vulnerabilities have focused on enhancing the robustness of classification models. 
In contrast, this paper adopts a fundamentally different strategy aimed at increasing 
robustness
through the redesign of road signs themselves. 
We propose an attacker-agnostic learning scheme to automatically design road signs that are robust to a wide array of patch-based attacks.
Empirical tests conducted in both digital and physical environments demonstrate that our approach significantly reduces vulnerability to patch attacks, outperforming existing techniques. 
\end{abstract}
\section{Introduction}

\begin{figure}[hb]
\centering
\includegraphics[width=0.90\textwidth]{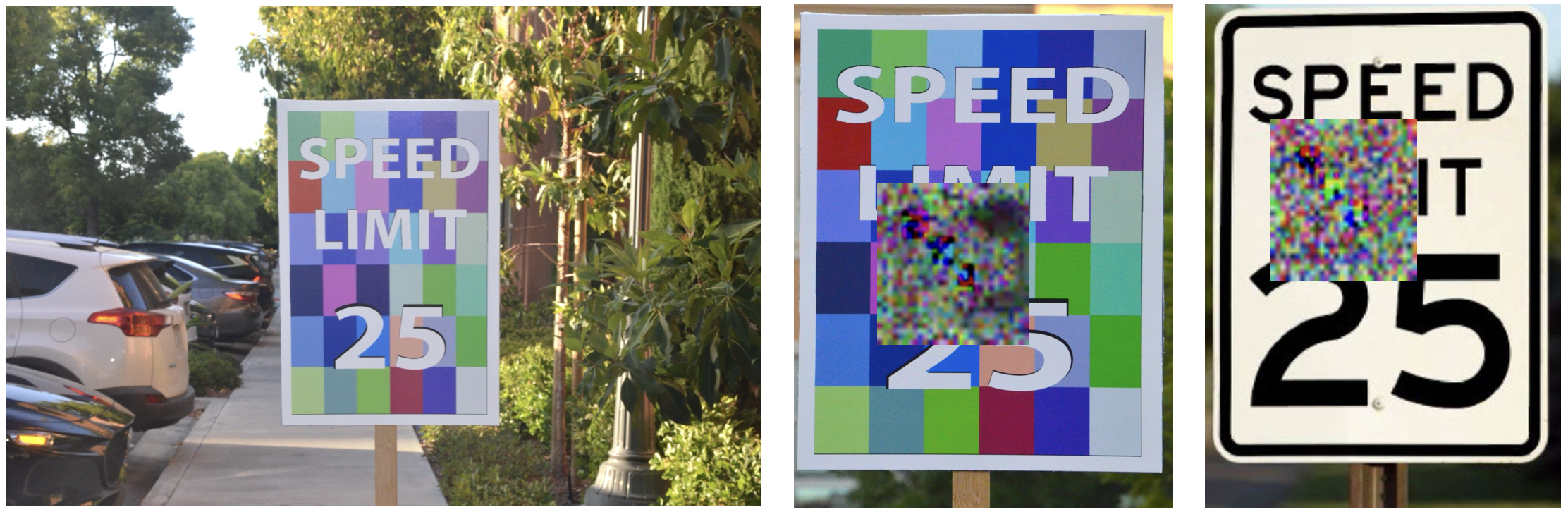}
\caption{Redesigned speed limit sign (left) with attacks on redesigned (middle) and original (right).
}
\label{fig:fig1}
\end{figure}

As autonomous driving systems become progressively more embedded in real-world systems, their safety becomes paramount. 
These systems and their sub-components, such as classification and segmentation modules, have been shown to be vulnerable to adversarial attacks \cite{goodfellow2014explaining,madry2017towards, kurakin2016adversarial} 
In this work, we focus on enhancing the safety of such systems by modifying the appearance of objects (specifically road signs) such that adversarial attacks applied to those objects are less effective (see Figure \ref{fig:fig1}).

\cite{salman2021unadversarial} first propose modifying the appearance of physical objects by designing patterns that make them easier to recognize under naturally challenging conditions, e.g., foggy weather. 
These conditions occur naturally rather than being the result of an adversarial attack.
Adversarially crafted perturbations pose a more significant challenge from a defender’s perspective for two key reasons: firstly, adversarial examples are explicitly designed to decrease model performance, and secondly, they are out of distribution with respect to training data (naturally challenging conditions are typically seen in training data, albeit scarcely for some domains).

In the context of autonomous driving, 
\cite{eykholt2018robust,yang2020patchattack} demonstrated the practical dangers of misclassification, such as small errors leading to severe accidents. 
Giving greater concern is the observation that these adversarial attacks can be physically realized \cite{eykholt2018robust, kurakin2016adversarial, athalye2018obfuscated}.
Physically realizable attacks often take the form of adversarial patches, small regions of an image designed to deceive classifiers, detectors, segmenators. 
\cite{brown2017adversarial,eykholt2018robust,liu2018delving,karmon2018lavan,zhang2019robust} introduced physical adversarial patches for real-world objects. 

Defenses can be categorized into \emph{Attack-Aware} and \emph{Attack-Agnostic} defenses. 
Attack-aware defenses, such as adversarial training,  (\cite{goodfellow2014explaining, madry2017towards, shafahi2019adversarial}), rely on knowledge of specific attacks.
 In contrast, \emph{Attack-Agnostic} defenses, such as randomized smoothing (\cite{cohen2019certified, lecuyer2019certified, salman2019provably}) and image sanitization techniques (\cite{xiang2020patchguard, xiang2022patchcleanser,xu2023patchzero}), do not rely on specific attack information.

To counter adversarial attacks, we propose an environmental-centric approach, Robust Environmental Design (\textbf{RED}), in which we design the backgrounds of road signs such that the road signs are both robust and still easy to print (as shown in Figure \ref{fig:fig1}). 
Our defense procedure and objective function learn object patterns so that once these road signs are deployed in the physical world, they achieve robustness against patch attacks \emph{without} requiring adversarial training. 
The model $f$ is trained only on clean data, and at test time, even some naive image sanitizing defense ensures $f$ can accurately predict even on maliciously modified inputs $x'$, eliminating the need for adversarial training after deployment. 
To demonstrate the efficacy of our method, we conduct experiments using two common benchmark datasets for road sign classification, LISA and GTSRB 
\cite{eykholt2018robust}, and test against several types of patch-based attack paradigms. 
Our approach achieves high levels of robustness compared to baseline models. 
Additionally, we conducted physical experiments by printing various common road signs (e.g., stop signs, speed limit signs, etc.) with patterns optimized via \textbf{RED}. We collected photos at different times of the day, under various lighting and weather conditions. We find that \textbf{RED} significantly improves robustness against attacks in both digital and physical settings.

\section{Preliminaries}

\textbf{Road Sign Classification}
Let $\X\subset \mathbb{R}^{W\times H}$ be a domain of possible road sign images of $W$ by $H$ pixels, and $Y$ be the set of possible road sign classes. Each road sign image $X\in\X$ has a corresponding true label $y\in Y$ (e.g., stop sign). 
To predict the class of a road sign, a classifier $f:\X \rightarrow Y$ is used, where $f(X)$ represents the predicted class of image $X$. 

\textbf{Adversarial Patch Attack}
We focus on patch-based attacks against image classification models. 
For a given image $X$ with true label $y$, the attacker's goal is to find a maliciously modified version of $X$, say \( X' \) such that \( f(X') \neq y \). 
Patch-based attacks constitute the attacker modifying a region of at most $B$ pixels in $X$.
The region can have various shapes but is constrained to be a contiguous region of the image, and is defined by a binary mask $M$ where $M[i][j] = 1$ if the attacker is modifying pixel $(i, j)$ and $M[i][j]=0$.
The attacker then applies a perturbation $\delta$ (with magnitude at most $\varepsilon$) to the pixels defined by $M$.
The attacker finds their desired mask $M$ and perturbation $\delta$ via the following:
\begin{align}
    \delta', M' =~& \arg\max_{M, \delta} \mathbb{P}\big(f(X') \neq y)\label{eq:adv}\\
                \text{s.t.} & ~\|\delta\|_{\infty} \leq \varepsilon\nonumber\\
                &|M| \leq B \nonumber\\
                & X' = (1-M) \odot X + M\odot \delta \nonumber
\end{align}
Where $|M|$ counts the number of 1's in the mask $M$ and $\odot$ is elementwise multiplication.



\section{RED-Robust Environmental Design}
Next, we outline our proposed method for robust environmental design (RED).
RED works by crafting the background of a road sign such that any patch placed on that road sign is not effective at fooling the classifier (see Figure \ref{fig:fig1} for an example).
RED has two key phases: a pattern selection phase in which we select the pattern for a given road sign class and a training phase in which we train our classifier on these newly selected patterns. 
The result of RED is a classifier that is robust to a wide array of patch-based attacks (without the need to simulate those attacks directly).

The key insight to our pattern selection is that road signs are manufactured objects, and their true label $y$ is known at manufacture time. 
Thus, we will seek to modify the road signs at manufacture time to contain a high level of \emph{class specific information}, making them easier to detect and, more importantly, harder to attack.
More formally, for each class $y$, let  \( \alpha_y \) represent the pattern on road signs of class $y$ (e.g., when $y$ is the class stop signs, the current design of $\alpha_y$ is a red background). 

We employ \emph{image ablation} defense at inference time, with the key advantage that no adversarial training is required to detect the adversarial patch. An ablation algorithm g masks image X, leaving only a subregion of unmasked pixels (see Figure \ref{fig:train}). Several works (\cite{xiang2020patchguard, xiang2021patchcleanser, levine2020randomized}) have explored this approach, differing mainly in ablation size and strategies for removing adversarial patches when specific attack information is available. Most of these defenses, however, are effective only for small patches, as they remove only a small portion of the image; when the patch covers more than 10\%, as shown in \cite{xiang2020patchguard, xiang2021patchcleanser}, they tend to fail, limiting their effectiveness as a universal defense.
In contrast, \cite{levine2020randomized} propose a technique that defends against patches of various sizes by applying a simple ablation technique that removes most of the image, preserving only a small portion for inference to avoid the patch. In this paper, we empirically demonstrate the robustness of redesigned road signs using this generic method \cite{levine2020randomized} to defend against both small and large adversarial patches. However, our road signs are not tied to any specific defense. In the extended version of this paper, we will present the effectiveness of these redesigned signs using various image sanitizing techniques like \cite{xiang2020patchguard} and \cite{xiang2022patchcleanser}.

At inference time, we apply several different ablation algorithms $g_1, \ldots g_m$, to image $X_{\alpha}$, producing masked images $g_1(X_{\alpha})\ldots g_m(X_{\alpha})$. 
The classifier then makes a prediction on each ablated image, and the final prediction is obtained via majority vote:
\[
    \text{majVote}\big(f\big(g_1(X_{\alpha})\big), \ldots, f\big(g_m(X_{\alpha})\big)\big)
\]
RED selects $\alpha_y$ such that $X_{\alpha_y}$ does contain enough class-specific information via Algorithm \ref{alg:red}.

\begin{algorithm}
\caption{Robust Environmental Design (RED)}
\label{alg:red}
\begin{algorithmic}[1]
\State \textbf{Input:} Dataset \(X, Y\), 
\State \textbf{Output:} Road sign backgrounds \(\boldsymbol{\alpha}\) for each class; \(\boldsymbol{\alpha} = \{\alpha_1, \ldots, \alpha_m \}\)
\State randomly initialize $\boldsymbol{\alpha}$
\For{each epoch}
    \State apply pattern $\alpha_y$ to each image $X$ with class $y$
    \State Compute the total loss $ \min_{\boldsymbol{\alpha}} \sum_{i=1}^N \sum_{j=1}^m L\bigg(f\big(g_j(X_{\alpha_{y_i}})\big), ~y_i\bigg)$
    \State Compute gradient of $\mathcal{L}$ w.r.t. to $\alpha$ and $f$, i.e. $\nabla_{\boldsymbol{\alpha}} L$ and $\nabla_{f} L$
    \State Update $\boldsymbol{\alpha}$ and $f$ according to $\nabla_{\boldsymbol{\alpha}}$ and $\nabla_{f}$ \textcolor{blue}{// In practice we parameterize $\alpha$ such that the resulting pattern in a checkerboard (as shown in Figure \ref{fig:train}) }
\EndFor
\State \textbf{Return} \(\boldsymbol{\alpha}\), $f$
\end{algorithmic}
\end{algorithm}

\textbf{Training}
Given any image ablation algorithm \( g \), a dataset with \( N \) examples, and \( N \) classes of road signs, our goal is to find a robust background $\alpha_y$ for each class $y$.
We use gradient methods to find $\alpha_y$; we parameterize the background of each sign to be a colored grid (see Figure \ref{fig:fig1}) where $\alpha_y$ gives the color of each element of the grid. 
We then minimize the classification loss with respect to color parameters $\alpha_1, \ldots, \alpha_N$, i.e., 
\begin{align}
&\min_{\boldsymbol{\alpha}} \sum_{i=1}^N \sum_{j=1}^m L\bigg(f\big(g_j(X_{\alpha_{y_i}})\big), ~y_i\bigg)\label{eq:min_alpha}
\end{align}
Figure \ref{fig:train} of the Appendix illustrates the training process.

\section{Experiments}\label{sec:exp}
\paragraph{Datasets and Road Sign Classification Models}
We conduct experiments on GTSRB and LISA datasets used in \cite{eykholt2018robust}.
GTSRB includes thousands of traffic signs across 43 categories of German road signs, while LISA contains 16 types of US road signs.
Following \cite{eykholt2018robust}, we used a simple CNN style with ReLU networks as the classification models for classifying each road sign dataset.  
For patch attacks, we utilize the \textbf{Sticker-Attack} defined by \cite{eykholt2018robust} and the \textbf{Patch-Attack} introduced by \cite{brown2017adversarial}.

\emph{Digital Redesigned Road Sign:}
to simulate the deployment of redesigned road signs, we applied spatial and color transformations to mimic real-world conditions, such as varying lighting and camera capture capabilities. For color changes, we sampled 22 contrast colors, printed them as patches, and photographed them under different lighting, distances, and times of day. 
For spatial transformations, we used bounding box annotations to learn homography matrices, simulating spatial changes in camera capture.
Given a top-down redesigned road sign,
We then simulate the deployment of the redesign by applying color and spatial transformations to the selected pattern, ensuring it fits onto the road sign. These digitally collected road signs will be referred to as digital-RED-LISA and digital-RED-GTSRB.

\emph{Physical Redesigned Road Signs:}
we printed the redesigned road signs and captured photos at various times of day, from different distances, and under varying lighting conditions between the road and the camera. Further details on the physical experiment setup are deferred to the corresponding section.
We will refer to these physically collected datasets as physical-RED-LISA and physical-RED-GTSRB. 

\paragraph{Robustness}
We are interested in defenses where the defender does not have access to the specifics of the attack, and the only prior knowledge is that the attack takes the form of a patch. 
In such defenses, the standard approach is to \emph{sanitize} the image before feeding it into the model \cite{levine2020randomized}.
In this short version paper, we use (De)Randomized defense for showing the robustness of RED designed signs.
In Table \ref{tab:lisa_results} , we show the performance under different attacks. (De)Randomized improves the model’s performance on adversarial examples compared to the vulnerability seen when the entire image is used as input. 
However, even with a patch attack covering only 10\% of the image, the efficacy of the road sign classification model drops from 82\% to 75\%.
Moreover, (De)Randomized exhibits decreased accuracy on clean data (compared to using the full image).
We observe that our method obtains significantly greater accuracy on both clean and attacked data compared to patch-smoothing.

\begin{table}[tbh]
  \label{tab:lisa_results}
  \centering
  \resizebox{0.85\textwidth}{!}{%
  \begin{tabular}{lcccc}
    \toprule
    \textbf{Method} & \textbf{Clean} & \textbf{Sticker Attack} & \textbf{Patch Attack (10\%)} & \textbf{Patch Attack (30\%)} \\ \midrule
    \textbf{LISA} & & &  & \\
    No Defense     & 99\% & 10\% & 5\%  & 10\% \\ 
    (De)Randomized & 82\% & 88\% & 75\% & 63\% \\ 
    RED \emph{(ours)} & 99\% & 99\% & 99\% & 95\% \\ \midrule
    \textbf{GTSRB} & & & & \\
    No Defense     & 99\% & 20\% & 15\% & 2\% \\ 
    (De)Randomized & 84\% & 75\% & 91\% & 87\% \\ 
    RED \emph{(ours)}           & 99\% & 99\% & 98\% & 99\% \\ \bottomrule
  \end{tabular}
  }
  \caption{Prediction accuracy on LISA, GTSRB and RED designed signs in LISA, GTSRB}
\end{table}

A key question we address is \emph{How to ensure a small patch contains enough information to accurately infer the class of an image}. We propose using a colorful grid as the background for road signs and conducted ablation analysis on grid size.
In Table \ref{tab:patch_accuracy} we show classification accuracy under different grid sizes for road sign background; S3, S5, S10 represent 3x3, 5x5, and 10x10 grid sizes (see Figure \ref{fig:pattern_size}). Note that the 1x1 grid is equivalent to the current road sign design. 
These results indicate that the current road sign designs do not always allow small patches to reliably classify the signs, but as we increase the grid size, even small defense mask sizes result in high accuracy (e.g., 90\% accuracy with mask size of only 13\%).
\begin{figure}[tbh]
\centering
\includegraphics[width=0.94\textwidth]{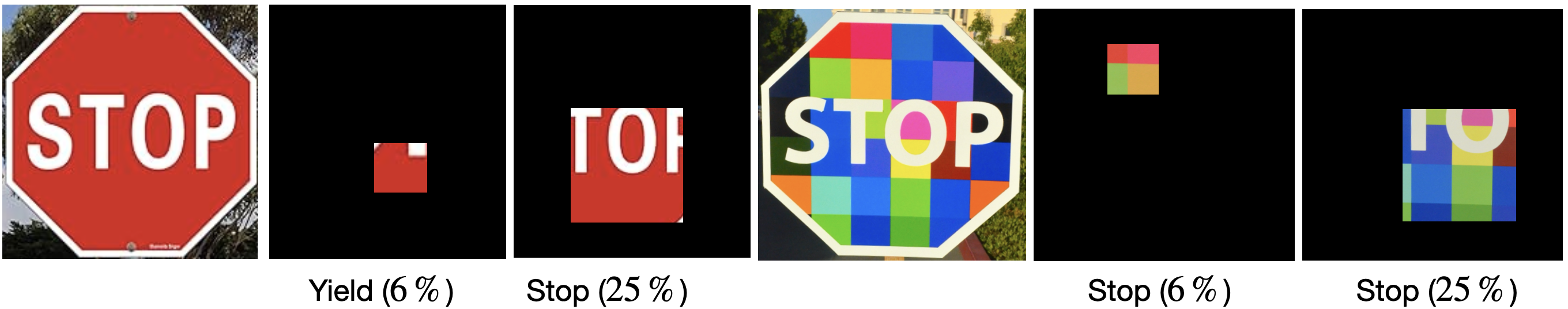}
\caption{Visualization of ablation sampling for LISA (left) and RED applied to LISA (right), with predicted class and ablation size percentage (bottom).}
\label{fig:albetion_sampling}
\end{figure}
This approach relies on ensuring that each sampled sign patch contains sufficient class-specific information to support independent inference of the road sign's class. 

\begin{table}[tbh]
  \caption{Ablation Analysis on Grid Size: Accuracy of defense mask across various grid sizes.}
  \label{tab:patch_accuracy}
  \centering
  \resizebox{0.95\textwidth}{!}{%
  \begin{tabular}{ccccccccc}
    \toprule
    \textbf{Defense Mask Size} & \textbf{GTSRB} & \textbf{GTSRB-S3} & \textbf{GTSRB-S5} & \textbf{GTSRB-S10} & \textbf{LISA} & \textbf{LISA-S3} & \textbf{LISA-S5} & \textbf{LISA-S10} \\ 
    \midrule
    13\%  & 48\% & 61\% & 55\% & 50\%    & 50\% & 96\% & 94\% & 90\%    \\ 
    20\% & 71\% & 98\% & 99\% & 88\%    & 65\% & 99\% & 99\% & 92\%    \\ 
    26\%  & 84\% & 99\% & 99\% & 99\%    & 78\% & 99\% & 99\% & 99\%    \\ 
    40\% & 95\% & 99\% & 99\% & 99\%    & 91\% & 99\% & 99\% & 99\%    \\ 
    \bottomrule
  \end{tabular}%
  }
\end{table}


\paragraph{Evaluation Across Different Attacks}
We evaluate our redesigned road signs against various attacks, including sticker attacks and patch attacks (PGD-inf) with different shapes: rectangles, triangles, and multi-patches. The multi-patch attack is specifically designed to target  ablated defense, where the attacker uses multiple small patches to bypass the defense. As shown in Table \ref{tab:results_1}, RED demonstrates strong performance against each variant of patch attack.

\noindent
\paragraph{Physical Experiment}
We physically recreated representative road signs for evaluation: two speed limit signs, a stop sign, and an arrow from LISA, and a stop sign, two speed limit signs, and a truck warning sign from GTSRB.

\begin{figure}[tbh]
\centering
\includegraphics[width=0.94\textwidth]{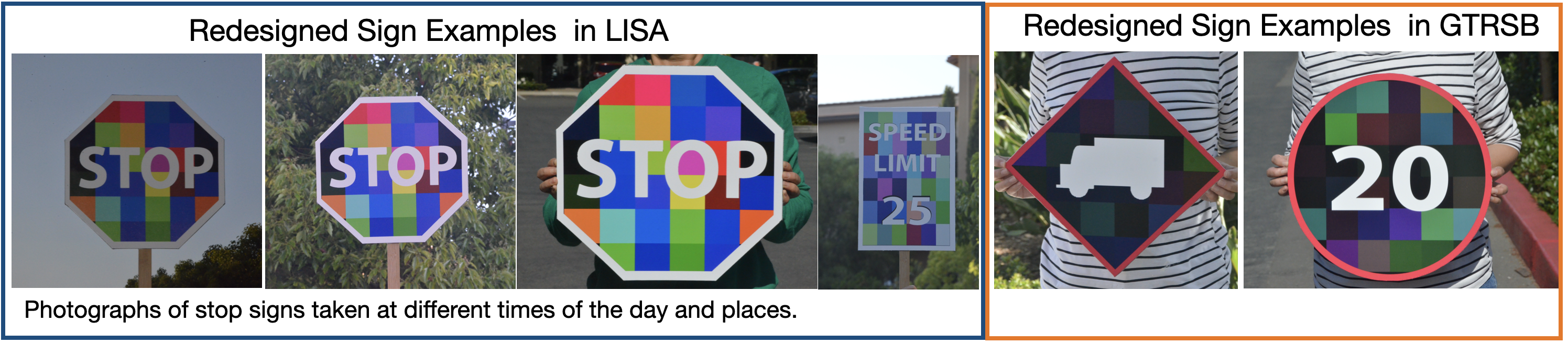}
\caption{Physical examples of patterns selected by RED.}
\label{figure:different_light}
\end{figure}

We show some examples in Figure \ref{figure:different_light}.
After finding redesigned road sign datasets for LISA and GTSRB via simulation, we printed them on 16x18 inch paper boards using a \emph{Sony Picture Station} printer. 
The signs were then photographed in various real-world settings using a Nikon D7000, either handheld or mounted on a wood stick (see Figure \ref{figure:different_light}). 
About 50 images of each sign were captured under diverse conditions, including different locations, weather, and times of day. 

Table \ref{tab:results_1} shows that our design is significantly more robust than current road signs when deployed in the physical world. Moreover, even with a simplified sanitizing algorithm \cite{levine2020randomized}, which uses only a small portion of the image for inference, the robust design—where each small area contains discriminative class information—achieves strong performance. The pattern produced by RED is resilient to both adversarial attacks and natural noise in the physical world.

\begin{table}[h]
  \caption{Evaluation of the Propose Methods Against Different Shape of Attack.}
  \label{tab:results_1}
  \centering
  \resizebox{\textwidth}{!}{%
  \begin{tabular}{cccccccc}
    \toprule
    \textbf{Datasets} & \textbf{LISA} & \textbf{RED-LISA-Digital} & \textbf{RED-LISA-Physical} & \textbf{GTSRB} & \textbf{RED-GTSRB-Digital} & \textbf{RED-GTSRB-Physical} \\
    \midrule
    Patch Size: \textbf{10\%}  & & & & & & \\
    Rectangle      & 75\% & 99\% & 99\% & 91\% & 99\% & 98\% \\
    Triangle       & 75\% & 99\% & 99\% & 91\% & 98\% & 99\% \\
    Multi-Patches  & 63\% & 99\% & 98\% & 83\% & 99\% & 97\% \\
    \midrule
    Patch Size: \textbf{30\%}  & & & & & & \\
    Rectangle      & 63\% & 94\% & 95\% & 87\% & 99\% & 99\% \\
    Triangle       & 63\% & 94\% & 95\% & 87\% & 94\% & 93\% \\
    Multi-Patches  & 59\% & 93\% & 95\% & 85\% & 93\% & 94\% \\
    \midrule
    Sticker        & 88\% & 99\% & 99\% & 75\%  &  99\% &  99\%  \\
    \bottomrule
  \end{tabular}%
  }
\end{table}

\clearpage
\section{Social Impact Statement}
Our work on Robust Environmental Design (RED) enhances the robustness of visual recognition systems, particularly for self-driving cars, by improving the resilience of road signs against adversarial attacks. This contributes to the safety and reliability of autonomous navigation technologies as they integrate into society.
The societal impact is significant, as our research reduces the risk of adversarial misclassifications, which could lead to traffic accidents or system failures. This directly improves public safety by strengthening the reliability of critical systems in urban and rural environments. 

Our work addresses the challenge of ensuring AI systems are both performant and resistant to manipulation. By developing defenses against adversarial attacks, we contribute to more secure and fair AI technology deployment. 
With this in mind, we note that although RED significantly improves robustness, it requires the ability to edit objects (e.g., selecting the pattern applied to road signs at manufacture time). This may not be feasible for all objects (e.g., pedestrians, wild animals, plants, etc.).

\bibliography{reference}

\clearpage
\appendix
\section*{Appendix}

\section{Methodology}
\paragraph{Enhance Class Information within a Road Sign}
In practice, we observe that smaller patch regions are more effective (see Section \ref{sec:exp} for a more thorough study of region size. Our findings across both the LISA and GTSRB datasets reveal that current sign designs typically require a relatively large visible area for effective inference. To address this issue, we propose redesigning road signs to enhance the informational content within small local areas, say small patches. 

Without loss of generality, we consider an ablation function \( g \), which obscures most of the image while retaining only a small patch. 
Consequently, an ablated sample \( s \) will contain just this small patch of the original image \( X \).
This approach serves as a showcase for the robust road sign design. 

We employ Algorithm \ref{alg:red} to optimize the design of robust road sign backgrounds. These backgrounds are engineered to enhance the class information within localized small areas. Consequently, as illustrated in Figure \ref{fig:albetion_sampling}, every local area of the newly designed road signs contains essential class information.
This redesigned strategy aims to ensure that even minimal patches can independently verify the sign's class, i.e., \( f\big(g(X_{\alpha})) = y \).

When selecting, the set of ablation functions $g_1\ldots g_m$, both the region and ablation size are consequential. 
Other works which use albetion function (e.g.,  \cite{xu2023patchzero}) suggest using a random size and location; in addition to one randomized abletion, we propose a majority vote-based algorithm to utilize \( S \) for inference. We will show the empirical results for both methods in the next section.

\paragraph{Training}
Each class has a pattern. For an image with label \( y_i \), the corresponding pattern is denoted as \( \alpha_{y_i} \). This pattern is then combined with the road sign mask, which includes text and shape, using precomputed color and homography mapping. 
The resulting image is processed, and the loss is calculated using Equation \ref{eq:min_alpha}. Finally, the gradients are backpropagated to update the parameters.

\begin{figure}[h]
\centering
\includegraphics[width=0.85\textwidth]{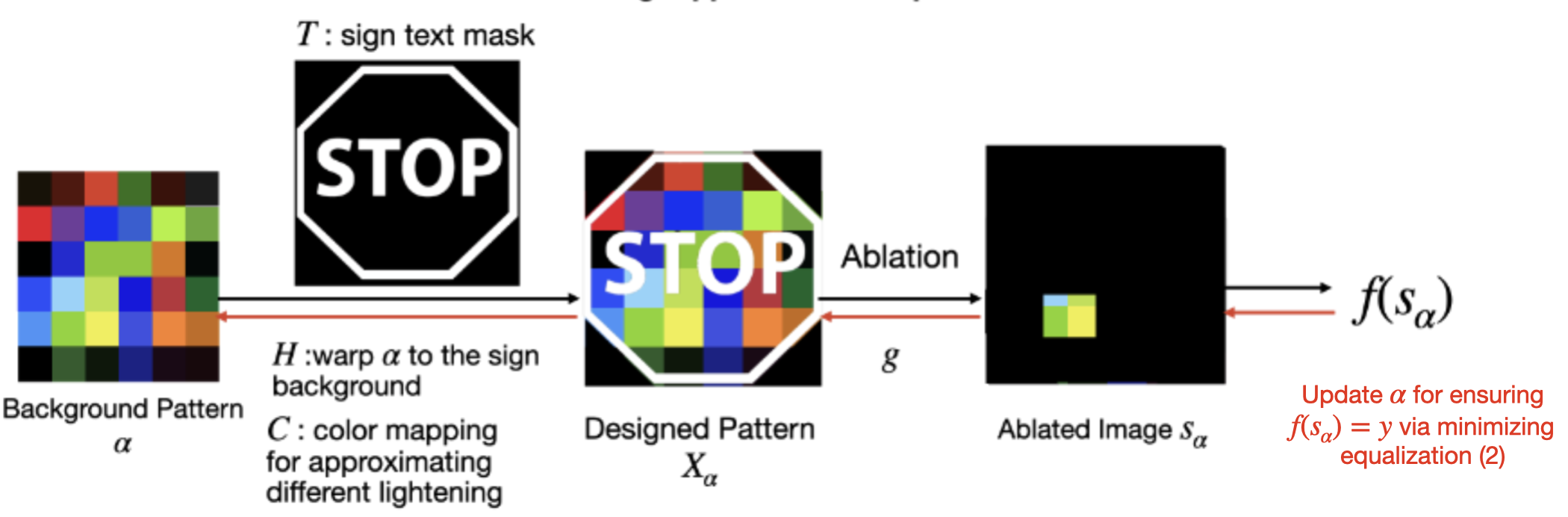}
\caption{Training Pipeline Via Differentiable Image Compositing}
\label{fig:train}
\end{figure}

Next, we will demonstrate an ablation algorithm \( g \) combined with our methods. We will discuss this in more detail.

\subsection{Special Case: Attacker-Aware Robust Environmental Design (AA-RED)}
Next, we look at how RED can be improved when the defender has knowledge of the attacks, and designs specific robust signs for robustness against given attacks $A$; the set of attacks is $\delta$.
Let $\alpha$ be the robust pattern, it is label-specific, and each class has a robust pattern, let $f$ be the classification model, and let $L$ be the cross entropy loss:
\begin{align*}
f^*, \boldsymbol{\alpha} = \min_{\alpha, f} \max_{\delta} &\sum_i \left( \underbrace{L\big(f\big(g(X_{\alpha})\big), y_i\big)}_{\text{loss on clean images}} + \underbrace{L\big(f\big(g(X_{\alpha}+\delta)\big), y_i\big)}_{\text{loss on adv images}} \right)\\
&\text{ s.t. }~\delta~\text{ is defined by Equation \ref{eq:adv}}
\label{eq:AA_loss}
\end{align*}
That is when the attacker is known, the defender can simulate the attacker's best response $\delta$ to the defender's current choice of pattern $\boldsymbol{\alpha}$ and classifier $f$.
This is effectively a combination of adversarial training and RED. 
The full procedure for AA-RED is outlined in Algorithm \ref{alg:optimize_patterns_with_attack}

\begin{algorithm}
\caption{Attacker-Aware Robust Environmental Desing (AA-RED)}
\label{alg:optimize_patterns_with_attack}
\begin{algorithmic}[1]
\State \textbf{Input:} Dataset \(X, Y\) 
\State \textbf{Output:} Road sign backgrounds \(\boldsymbol{\alpha}\) for each class; \(\boldsymbol{\alpha} = \{\alpha_1, \cdots, \alpha_j, \ldots, \alpha_m \}\)
\State randomly initialize $\boldsymbol{\alpha}$
\For{each epoch}
    \State apply pattern $\alpha_y$ to each image $X$ with class $y$
    \State Generate ablated images using \(g\)
    \State Compute the attacker's best perturbation $\delta_i$ for each each modified image $X_{i, \alpha_{y_i}}$
    \State Compute the total loss $ \mathcal{L} = \sum_{i=1}^N \bigg(L\big(f\big(g(X_{i, \alpha})\big), y_i\big) 
 +L\big(f\big(g(X_{i, \alpha}+\delta)\big), y_i\big)\bigg)$
    \State Compute gradient of $\mathcal{L}$ w.r.t. to $\alpha$ and $f$, i.e. $\nabla_{\boldsymbol{\alpha}} L$ and $\nabla_{f} L$
    \State Update $\boldsymbol{\alpha}$ and $f$ according to $\nabla_{\boldsymbol{\alpha}}$ and $\nabla_{f}$
\EndFor
\State \textbf{Return} \(\boldsymbol{\alpha}\), $f$
\end{algorithmic}
\end{algorithm}

\begin{algorithm}
\caption{Inference Algorithm (Majority Vote)}
\label{alg:majority_vote}
\begin{algorithmic}[1]
\State \textbf{Input:} Image $X$, ablation function $g$, model $f$
\State \textbf{Output:} Prediction for $X'$
\State predictions $ = \emptyset$
\For{$j=1 \ldots m$ }
    \State $p = f\big(g_j(X)\big)$ \textcolor{blue}{// Prediction for the $j^{(\text{th}}$ ablution of $X$}
    \State Predictions.add$(p)$
\EndFor
\State finalPrediction $ = $ mode$($predictions$)$
\State \textbf{Return} finalPrediction
\end{algorithmic}
\end{algorithm}

\begin{table}[tbh]
  \label{tab:lisa_results_full}
  \centering
  \resizebox{\textwidth}{!}{%
  \begin{tabular}{lcccccc}
    \toprule
    \textbf{Method} & \textbf{Clean} & \textbf{Sticker Attack} & \textbf{Patch Attack (5\%)} & \textbf{Patch Attack (10\%)} & \textbf{Patch Attack (25\%)} & \textbf{Patch Attack (30\%)} \\ \midrule
    \textbf{LISA} & & & & & \\
    No Defense & 99\% & 10\% & 15\% & 5\% & 10\% & 10\% \\ 
    (De)Randomized & 82\% & 88\% & 81\% & 75\% & 67\% & 63\% \\ 
    RED      & 99\% & 99\% & 99\% & 99\% & 96\% & 95\% \\ \midrule
    \textbf{GTSRB} & & & & & \\
    No Defense  & 99\% & 20\% & 22\% & 15\% & 8\% & 2\% \\ 
    (De)Randomized & 96\% & 75\% & 92\% & 91\% & 90\% & 87\% \\ 
    RED      & 99\% & 99\% & 99\% & 98\% & 98\% & 99\% \\ \bottomrule
  \end{tabular}
  }
  \caption{Prediction accuracy on LISA, GTSRB and \textbf{RED} designed signs in LISA, GTSRB}
  
\end{table}

\section{Experiments}
In this section, we present RED designed LISA and GTSRB results against a wider range of attacks, which could not be included in the main body due to space constraints.
\begin{table}[htbp]
  \caption{Evalution of the Propose Methods Against Different Shape of Attack.}
  \label{tab:results}
  \centering
  \resizebox{\textwidth}{!}{%
  \begin{tabular}{|c|ccc|ccc|}
    \toprule
    \textbf{Datasets} & \textbf{LISA} & \textbf{RED-LISA-Digital} & \textbf{RED-LISA-Physical} & \textbf{GTSRB} & \textbf{RED-GTSRB-Digital} & \textbf{RED-GTSRB-Physical} \\
    \midrule
    Patch Size: \textbf{5\%} & & & & & & \\
    Rectangle      & 81\% & 99\% & 99\% & 93\% &  99\% &  99\%  \\
    Triangle       & 80\% & 99\% & 99\% & 92\% &  99\% &  98\% \\
    Multi-Patches  & 82\% & 99\% & 98\% & 88\% &  99\% &  99\%  \\ 
    \midrule
    Patch Size: \textbf{10\%}  & & & & & & \\
    Rectangle      & 75\% & 99\% & 99\% & 91\% & 99\% & 98\% \\
    Triangle       & 75\% & 99\% & 99\% & 91\% & 98\% & 99\% \\
    Multi-Patches  & 63\% & 99\% & 98\% & 83\% & 99\% & 97\% \\
    \midrule
    Patch Size: \textbf{25\%}  & & & & & & \\
    Rectangle      & 67\% & 97\% & 96\% & 90\% & 99\% & 98\% \\
    Trianlge       & 63\% & 94\% & 95\% & 87\% & 98\% & 99\% \\
    Multi-Patches  & 59\% & 95\% & 96\% & 71\% & 98\% & 98\% \\
    \midrule
    Patch Size: \textbf{30\%}  & & & & & & \\
    Rectangle      & 63\% & 94\% & 95\% & 87\% & 99\% & 99\% \\
    Triangle       & 63\% & 94\% & 95\% & 87\% & 94\% & 93\% \\
    
    Multi-Patches  & 59\% & 93\% & 95\% & 85\% & 93\% & 94\% \\
    \midrule
    Sticker        & 88\% & 99\% & 99\% & 75\%  &  99\% &  99\%  \\
    \bottomrule
  \end{tabular}%
  }
\end{table}

\begin{figure}[tbh]
\centering
\includegraphics[width=0.94\textwidth]{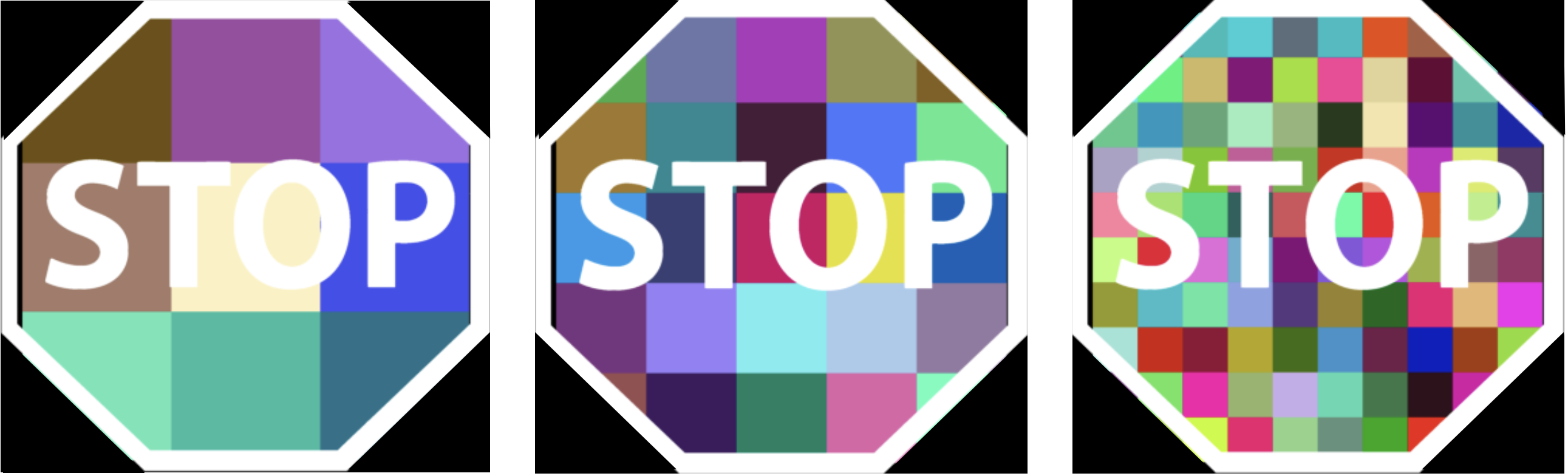}
\caption{Visualization of road signs with different grid pattern sizes: left (grid size 3), middle (grid size 5), and right (grid size 10)}
\label{fig:pattern_size}
\end{figure}

\end{document}